\documentclass[letterpaper]{article} 
\usepackage{aaai2026}  
\usepackage{times}  
\usepackage{helvet}  
\usepackage{courier}  
\usepackage[hyphens]{url}  
\usepackage{graphicx} 
\usepackage{natbib}  
\usepackage{caption} 
\usepackage{algorithm}
\usepackage{algorithmic}
\usepackage{amsmath,xcolor,amssymb}
\usepackage{booktabs}
\usepackage{multirow}
\usepackage{hhline}
\usepackage{bm}
\usepackage{arydshln}
\usepackage{tabularray}
\UseTblrLibrary{booktabs}
\usepackage{newfloat}
\usepackage{listings}
\DeclareCaptionStyle{ruled}{labelfont=normalfont,labelsep=colon,strut=off} 
\floatstyle{ruled}
\newfloat{listing}{tb}{lst}{}
\floatname{listing}{Listing}
\title{FIA-Edit: Frequency-Interactive Attention for Efficient and High-Fidelity Inversion-Free Text-Guided Image Editing}
\author{
    Kaixiang Yang\textsuperscript{\rm 1, 2, ${\dag}$}, Boyang Shen\textsuperscript{\rm 1, 2, ${\dag}$}, Xin Li\textsuperscript{\rm 1, 2}, Yuchen Dai\textsuperscript{\rm 2}, Yuxuan Luo\textsuperscript{\rm 2}, Yueran Ma\textsuperscript{\rm 2}\\ 
    Wei Fang\textsuperscript{\rm 3}, Qiang Li\textsuperscript{\rm 1, *}, Zhiwei Wang\textsuperscript{\rm 1, *}\\
}
\affiliations{
    \textsuperscript{\rm 1}Wuhan National Laboratory for Optoelectronics, Huazhong University of Science and Technology\\
    \textsuperscript{\rm 2}College of Life Science and Technology, Huazhong University of Science and Technology\\
    \textsuperscript{\rm 3}Wuhan United Imaging Healthcare Surgical Technology Co., Ltd\\
    $^\dag$: Co-first authors, $^*$: Corresponding authors.\\
    \{kxyang, boyangshen, liqiang8, zwwang\}@hust.edu.cn
}

\usepackage{bibentry}

\begin{document}

\maketitle

\begin{abstract}
Text-guided image editing has advanced rapidly with the rise of diffusion models. While flow-based inversion-free methods offer high efficiency by avoiding latent inversion, they often fail to effectively integrate source information, leading to poor background preservation, spatial inconsistencies, and over-editing due to the lack of effective integration of source information.
In this paper, we present \textbf{FIA-Edit}, a novel inversion-free framework that achieves high-fidelity and semantically precise edits through a \textbf{F}requency-\textbf{I}nteractive \textbf{A}ttention. Specifically, we design two key components: (1) a Frequency Representation Interaction (FRI) module that enhances cross-domain alignment by exchanging frequency components between source and target features within self-attention, and (2) a Feature Injection (FIJ) module that explicitly incorporates source-side queries, keys, values, and text embeddings into the target branch's cross-attention to preserve structure and semantics. 
Comprehensive and extensive experiments demonstrate that FIA-Edit supports high-fidelity editing at low computational cost ($\sim$6s per $512\times 512$ image on an RTX 4090) and consistently outperforms existing methods across diverse tasks in visual quality, background fidelity, and controllability.
Furthermore, we are the first to extend text-guided image editing to clinical applications. By synthesizing anatomically coherent hemorrhage variations in surgical images, FIA-Edit opens new opportunities for medical data augmentation and delivers significant gains in downstream bleeding classification. Our project is available at: https://github.com/kk42yy/FIA-Edit.

\end{abstract}


\begin{figure}[ht]
\begin{center}
	\includegraphics[width=0.9\linewidth]{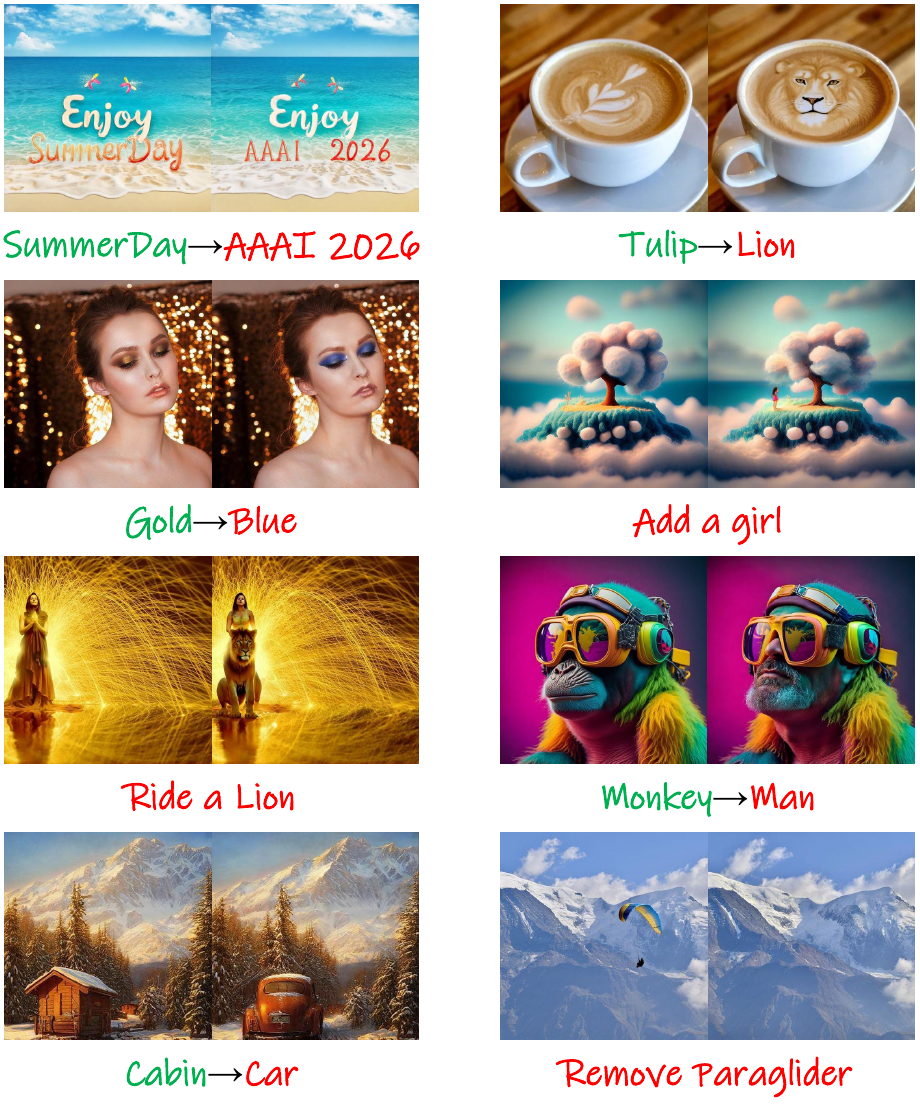}
\end{center}
\caption{
FIA-Edit is capable of handling a wide range of image editing tasks, including object modification, addition and removal, color transformation, and text replacement.
}
\label{fig:abs}
\end{figure}

\section{Introduction}

Text-guided image editing aims to modify an image according to a given textual description while preserving content unrelated to the edit. This task has witnessed significant progress in recent years, driven by the development of powerful generative models such as Denoising Diffusion Probabilistic Models (DDPMs)~\cite{DDPM,DDIM}, Latent Diffusion Models (LDMs)~\cite{LDM1.5}, and Diffusion Transformers (DiTs)~\cite{DiT,flux2024}. These models have been widely applied in real-world scenarios containing video editing~\cite{VideoEdit1,VideoEdit2,VideoEdit3,VideoEdit4,VideoEdit5,VideoEdit6}, visual effects production, and social media content creation.

Among these advances, tuning-free diffusion-based methods (\textit{e.g.}, DDIM-based sampling~\cite{P2P,PnP,MasaCtrl,FlexiEdit,FreeDiff,InsP2P} and Rectified Flow~\cite{StableFlow,Fireflow,RF-Inv,Rf-solver,DCEdit}) have gained increasing attention. These approaches eliminate the need for per-instance fine-tuning, allowing for flexible and efficient zero-shot editing. Most existing methods fall into one of two categories, each presenting a fundamental trade-off between editing fidelity and computational efficiency.

\begin{figure}
\begin{center}
	\includegraphics[width=1\linewidth]{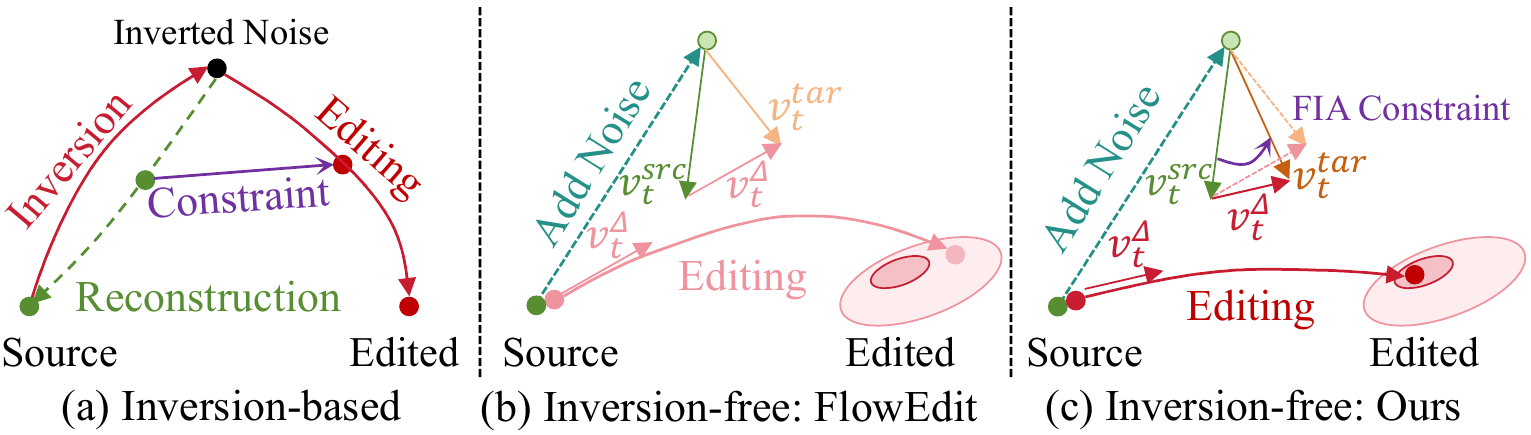}
\end{center}
\caption{
Overview of inversion-based and inversion-free image editing methods.
(a) Inversion-based methods first invert the source image to noise, then edit from noise using the target prompt, often injecting source features during denoising.
(b) Inversion-free methods bypass inversion by estimating velocity fields from noisy latent to source and noisy latent to target. Their difference defines the editing direction from source to target. However, this does not guarantee that the result preserves both background and target semantics (dark red ellipse).
(c) Our method incorporates source-aware constraints (\textit{i.e.}, FIA Constraint) during the computation of the noisy latent to target velocity field $v^{tar}_t$, effectively guiding the editing trajectory toward regions that preserve background fidelity while achieving semantic accuracy. Dashed arrows in (c) indicate FlowEdit, which lacks this guidance and fails to reach such optimal region.
}
\label{fig:overview}
\end{figure}

The first and more established category adopts an \emph{inversion-first} paradigm~\cite{P2P,PnP,MasaCtrl,FreeDiff}, where the source image is first projected into a latent prior distribution, typically Gaussian noise, using techniques such as DDIM inversion~\cite{DDIM}, Rectified Flow inversion~\cite{Flow1,Flow2,Flow3}, or more advanced schemes~\cite{NegPmt,NullPmt}. 
The editing process is then carried out in two stages: reconstructing the source image from the latent using the source prompt, and navigating from the same latent point to the target using the edited prompt. In this design, the prior distribution serves as a central ``waypoint'' that connects the source and target domains. To better align content and structure across the two branches, various feature interaction strategies including attention replacement and prompt injection have been proposed (see Fig.~\ref{fig:overview}a). However, the inversion step is computationally intensive and significantly slows down the overall editing pipeline.

To improve efficiency, recent works have explored \emph{inversion-free} approaches~\cite{InfEdit,FlowEdit,FlowAlign}. These methods avoid explicit mapping to the latent prior and instead aim to directly construct the source-to-target trajectory.
Since the source-to-target path is not directly accessible, these methods introduce virtual intermediate states by injecting noise into the source image. 
From this noisy reference point, two velocity fields are estimated: one pointing back to the source using source-prompt and the other toward the target using target-prompt. 
The vector difference between these two flows serves as an approximation of the editing direction, which implicitly encodes the source-to-target semantic transformation without ever performing an actual inversion.
While this design enables much faster inference, it lacks explicit integration of source features during the editing process. As shown in Fig.~\ref{fig:overview}b, this often results in poor content preservation in non-edited regions, leading to semantic drift, spatial inconsistency, and over-editing artifacts.

To address these limitations, we propose \textbf{FIA-Edit} (Fig.~\ref{fig:overview}c), a novel inversion-free image editing framework that achieves high editing quality, strong background preservation, and fast generation. Instead of relying on an implicit feature transformation, FIA-Edit introduces an \emph{explicit feature-level interaction} mechanism between the source and target representations throughout the editing trajectory. This design improves structural consistency and mitigates semantic drift in background regions.

The core of our method lies in a lightweight \textit{Frequency-Interactive Attention} architecture. It contains two key modules: (1) The Frequency Representation Interaction (FRI) module, which fuses source and target features in the frequency domain within self-attention blocks, promoting cross-domain alignment without additional memory cost; (2) The Feature Injection (FIJ) module, which injects source-side queries, keys, values, and text embeddings into the cross-attention layers of the target branch, enhancing spatial and semantic consistency.

Our main contributions are summarized as follows:
\begin{itemize}
    \item We propose \textbf{FIA-Edit}, an efficient and inversion-free image editing framework that achieves high-fidelity edits while explicitly preserving background structures.
    \item We introduce a unified Frequency-Interactive Attention mechanism, consisting of the FRI and FIJ modules, which enable explicit feature-level interaction between source and target to improve content alignment and structural consistency.
    \item We conduct extensive experiments on the PIE-Bench benchmark, and demonstrate that FIA-Edit achieves state-of-the-art performance across diverse editing tasks, with superior background preservation and semantic controllability.
    \item To the best of our knowledge, we are the first to apply general-purpose text-guided image editing methods to clinical images. Moving beyond artistic manipulation, FIA-Edit enables anatomically meaningful modifications, such as adjusting bleeding severity in surgical scenes. This opens up new opportunities for using image editing tools in medical data augmentation and downstream clinical tasks.
\end{itemize}

\section{Related Works}
\subsection{Inversion-based Methods}

Inversion-based image editing methods typically rely on first inverting the source image back into noise through an inversion process, and then performing editing conditioned on target prompts. Broadly, existing work in this category can be divided into three main directions:

\textbf{Improvements to the inversion process.}
A number of approaches aim to enhance the quality, stability, and accuracy of the inversion process. In the DDIM-based setting, 
Null-Text Inversion~\cite{NullPmt} demonstrates that effective inversion can be achieved without any textual prompt to suppress irrelevant content during reconstruction.
For rectified flow–based approaches, methods such as RF-Inv~\cite{RF-Inv} and FireFlow~\cite{Fireflow} focus on refining the inversion process to reduce reconstruction artifacts. Other methods, including Direct Inversion~\cite{PIE-Bench-DirInvPmt} and DNAEdit~\cite{DNAEdit}, aim to minimize the discrepancy between the actual and ideal inversion outputs, thereby boosting the quality of reconstruction.

\textbf{Feature injection during editing.}
To improve controllability and fidelity, many tuning-free approaches incorporate feature injection mechanisms during the generation process.
Prompt-to-Prompt (P2P)~\cite{P2P} explores direct feature replacement within cross-attention layers, whereas Plug-and-Play (PnP)~\cite{PnP} injects source features between residual and attention blocks to enhance background preservation.
FTEdit~\cite{FTEdit} introduces semantic feature replacement within adaptive layer normalization modules, enabling more precise and disentangled control over the generated content.

\textbf{Frequency-aware latent processing.}
Recent studies have increasingly explored the integration of frequency operations.
FlexiEdit~\cite{FlexiEdit} suppresses high-frequency components in DDIM latents associated with editable regions, enabling non-rigid edits.
FDS~\cite{FDS} adopts wavelet decomposition to adaptively select frequency bands according to the editing task, enabling fine-grained control. 
Despite their effectiveness, these methods require an inversion process, increasing editing time and involving multiple task-specific hyperparameters.

\subsection{Inversion-free Methods}
To reduce the computational overhead of image editing, a natural direction is to eliminate the time-consuming inversion process. Several recent approaches have explored this idea by bypassing the explicit mapping of source images into the noise space.
InfEdit~\cite{InfEdit} introduces the Denoising Diffusion Consistent Model (DDCM), which adopts a multi-step consistency sampling strategy that enables image editing without requiring explicit inversion.
FlowEdit~\cite{FlowEdit} further proposes an inversion-free framework by leveraging the velocity field to construct a direct trajectory from the source image to the edited target, avoiding inversion to Gaussian noise.
Building on this, FlowAlign~\cite{FlowAlign} introduces trajectory regularization to achieve more consistent and controllable text-driven editing within this inversion-free paradigm.

While these methods significantly reduce editing time, they often underutilize source image features, leading to insufficient background preservation and noticeable inconsistencies in non-edited regions.
In contrast, we propose FIA-Edit, which enhances inversion-free editing by introducing frequency-aware feature interaction directly within the velocity field. This design effectively retains high-fidelity background information while ensuring both editing quality and runtime efficiency.

\begin{figure*}[ht]
\begin{center}
	\includegraphics[width=0.8\linewidth]{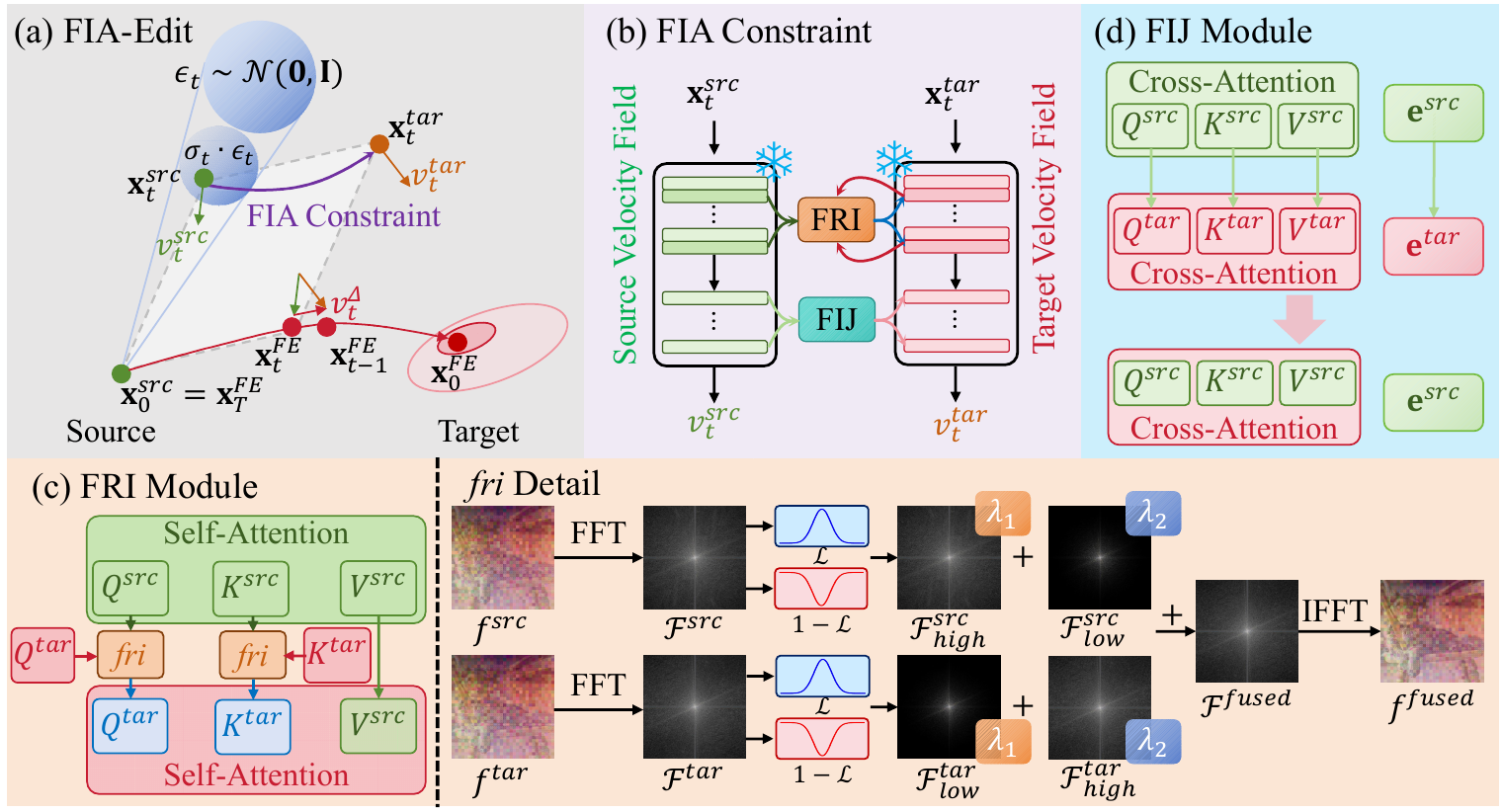}
\end{center}
\caption{
Details of our framework.
(a) Overview of FIA-Edit. During the computation of source and target velocity fields, we introduce the FIA constraint to enable interaction between source and target features. 
(b) FIA constraint.
(c) Frequency Representation Interaction (FRI). FRI is integrated into the self-attention layers. Both source and target Q/K features are fused in the frequency domain, and the fused output replaces the target Q/K. The right side shows the detailed structure of the frequency-domain fusion module $fri$.
(d) Feature Injection (FIJ). FIJ is used in the cross-attention layers in the latter of DiT.
}
\label{fig:FRI-FIJ}
\end{figure*}

\section{Method}
Fig.~\ref{fig:FRI-FIJ} illustrates the overall architecture of our method. As shown in Fig.~\ref{fig:FRI-FIJ}a, our approach builds upon the inversion-free FlowEdit paradigm~\cite{FlowEdit} as backbone. On this basis, we enhance the integration of relevant information from the source velocity field into the target velocity field, thereby improving both background preservation and semantic consistency during editing. Then, we detail the backbone and the proposed FIA Constraint, which consists of two sub-modules: Frequency Representation Interaction (FRI) and Feature Injection (FIJ).

\subsection{Backbone}
Our backbone is built upon Rectified Flow, which enables direct progression from the source domain to the target domain by estimating the difference between their respective velocity fields.

At a given discrete editing time step $\sigma_t$ with index $t$, a linear interpolation is first employed between the source image $\mathbf{X}^{src}$ and Gaussian noise $\mathcal{N}(\mathbf{0}, \mathbf{I})$, following the Rectified Flow formulation:
\begin{equation}
    \mathbf{x}^{src}_t = (1 - \sigma_t)\cdot \mathbf{X}^{src} + \sigma_t\cdot \epsilon_t, \quad \epsilon_t \sim \mathcal{N}(\mathbf{0}, \mathbf{I}).
\end{equation}
Using the source prompt $\mathcal{P}^{src}$, corresponding source velocity $v_\theta(\mathbf{x}^{src}_t, \mathcal{P}^{src}, t)$ is computed.

Next, to obtain the corresponding target velocity field, the current editing feature $\mathbf{x}^{FE}_t$ and the additive relationship of vectors $\mathbf{x}^{FE} = \mathbf{X}^{src} + \mathbf{x}^{tar} - \mathbf{x}^{src}$ are leveraged. Accordingly, the target representation at step $t$ can be expressed as:
\begin{equation}
    \mathbf{x}^{tar}_t = \mathbf{x}^{FE}_t + \mathbf{x}^{src}_t - \mathbf{X}^{src}.
\end{equation}
Target velocity $v_\theta(\mathbf{x}^{tar}_t, \mathcal{P}^{tar}, t)$ is obtained with target prompt $\mathcal{P}^{tar}$.
It is worth noting that during the editing process, $\mathbf{x}^{src}_t$ and $\mathbf{x}^{tar}_t$ progressively move toward the source and target domains, respectively.
At the initial step, the two are identical.

The direction is then determined by velocity difference:
\begin{equation}
    v^{\Delta}_t = v_\theta(\mathbf{x}^{tar}_t, \mathcal{P}^{tar}, t) - v_\theta(\mathbf{x}^{src}_t, \mathcal{P}^{src}, t).
\label{eq:delta_v}
\end{equation}
As evident, the backbone relies solely on the implicit interaction between $\mathbf{x}^{src}_t$ and $\mathbf{x}^{tar}_t$, without any explicit guidance from the source image features. This often causes the editing process to deviate toward the target domain too freely, leading to weak constraints from $\mathbf{X}^{src}$ and suboptimal results, especially in preserving source-relevant content.

Finally, the editing feature is updated iteratively according to the rectified flow stepping rule:
\begin{equation}
    \mathbf{x}^{FE}_{t-1} = \mathbf{x}^{FE}_t + (\sigma_{t-1} - \sigma_t)\cdot v^{\Delta}_t. 
\end{equation}
After completing all time steps, the edited image is synthesized from the final state $\mathbf{x}^{FE}_0$.

\subsection{FIA Constraint}
As shown in Fig.~\ref{fig:FRI-FIJ}b, to better preserve background content and ensure semantic alignment, we explicitly incorporate source features into the computation of target velocity fields through a proposed \textbf{FIA Constraint}. This constraint consists of two key components:
(1) the FRI module, which operates within \textit{self-attention} to enable frequency-domain interaction between source and target features, and
(2) the FIJ module, applied in \textit{cross-attention} to inject source features directly.

We denote the attention features extracted from $ v_\theta(\mathbf{x}^{src}_t, \mathcal{P}^{src}, t)$ and $ v_\theta(\mathbf{x}^{tar}_t, \mathcal{P}^{tar}, t)$ as $\{f^{src}_t\}$ and $\{f^{tar}_t\}$, respectively.
Then, Eq.~\ref{eq:delta_v} can be reformulated as:
\begin{equation}
    \begin{aligned}
    v^{\Delta}_t &= v_\theta(\mathbf{x}^{tar}_t, \mathcal{P}^{tar}, t, \mathtt{FIA}(\{f^{src}_t\},\{f^{tar}_t\})) \\ &- v_\theta(\mathbf{x}^{src}_t, \mathcal{P}^{src}, t)
    \end{aligned}
\end{equation}

\subsection{Frequency Representation Interaction}

To preserve structural fidelity while enabling meaningful semantic transformation, we introduce the Frequency Representation Interaction (FRI) module, which performs cross-domain feature fusion in the frequency domain, as illustrated in Fig.~\ref{fig:FRI-FIJ}c.

FRI is motivated by the observation that structure and semantics are more naturally disentangled in the frequency space: low-frequency components primarily encode coarse spatial layouts and background structures, while high-frequency components capture fine-grained textures and semantic details.
Based on this insight, we propose a cross-domain frequency fusion strategy that enhances the high-frequency components of the source and the low-frequency components of the target, while suppressing low-frequency content in the source and high-frequency signals in the target. This selective fusion effectively leverages source information through frequency-domain interaction.

We first compute the velocity fields of $\mathbf{x}^{src}_t$ and $\mathbf{x}^{tar}_t$, and extract the intermediate features $f^{src}_t \in \mathbb{R}^{C \times H \times W}$ and $f^{tar}_t \in \mathbb{R}^{C \times H \times W}$ from DiT. A 2D Fast Fourier Transform (FFT) $\mathtt{FFT(\cdot)}$ is then applied:
\begin{equation}
\mathcal{F}^{src} = \mathtt{FFT}(f^{src}_t), \quad \mathcal{F}^{tar} = \mathtt{FFT}(f^{tar}_t).
\end{equation}
Using a Gaussian low-pass filter $\mathcal{L}$, we decompose each into high- and low-frequency components:
\begin{align}
\mathcal{F}^{src}_{high} &= \mathcal{F}^{src} \cdot (1 - \mathcal{L}), \quad \mathcal{F}^{src}_{low} = \mathcal{F}^{src} \cdot \mathcal{L}, \\
\mathcal{F}^{tar}_{high} &= \mathcal{F}^{tar} \cdot (1 - \mathcal{L}), \quad \mathcal{F}^{tar}_{low} = \mathcal{F}^{tar} \cdot \mathcal{L}.
\end{align}

The fused spectrum is computed by applying cross-weighted fusion:
\begin{equation}
\mathcal{F}^{{fused}} = \lambda_1 \cdot (\mathcal{F}^{src}_{high} + \mathcal{F}^{tar}_{low}) + \lambda_2 \cdot (\mathcal{F}^{src}_{low} + \mathcal{F}^{tar}_{high}),
\end{equation}
where $\lambda_1 = 0.8$, $\lambda_2 = 0.2$ are weighting coefficients, emphasizing structure and semantics from source image while suppressing conflicting signals.

The fused feature is then reconstructed via inverse FFT:
\begin{equation}
f^{fused} = \mathtt{IFFT}(\mathcal{F}^{{fused}}),
\end{equation}
which is then injected into the self-attention layers, guiding the target velocity updates. By aligning complementary information across domains, FRI allows our model to perform semantically accurate edits while preserving the source's visual structure, improving both realism and control.

\subsection{Feature Injection}
To further improve background preservation, we draw inspiration from inversion-based methods~\cite{PnP,MasaCtrl} that inject source features to maintain spatial consistency and fine-grained control. We propose a Feature Injection (FIJ) module (Fig.~\ref{fig:FRI-FIJ}d) to explicitly introduce source features into the editing process.

Unlike prior works that inject only Q or K across the entire network, FIJ operates within the cross-attention layers of the later DiT blocks (\textit{i.e.}, layers $\mathtt{13\sim23}$). Specifically, we inject source-side query ($Q^{src}$), key ($K^{src}$), value ($V^{src}$), and text embedding ($\mathbf{e}^{src}$) into the target attention computation:
\begin{equation}
Q^{tar} \leftarrow Q^{src}, K^{tar} \leftarrow K^{src}, V^{tar} \leftarrow V^{src}, \mathbf{e}^{tar} \leftarrow \mathbf{e}^{src}.
\end{equation}
The injection is applied only during the early generation steps, when $\mathbf{x}^{tar}_t$ and $\mathbf{x}^{src}_t$ are still similar. This early fusion allows $\mathbf{x}^{tar}_t$ to absorb source information smoothly under $\mathcal{P}^{tar}$’s guidance, enabling coherent edits and avoiding the abrupt changes seen in other methods. Additionally, FIJ stabilizes the semantic alignment between source and target prompts.
\section{Experiment}
\subsection{Experiment Design}
\textbf{Dataset and Baselines.}
To thoroughly evaluate the effectiveness of our proposed method, we conduct experiments on the PIE-Bench~\cite{PIE-Bench-DirInvPmt} benchmark, which comprises 700 image–prompt pairs spanning 10 diverse editing categories.
We compare our approach with a comprehensive set of baselines, including:
LDM-based methods (P2P~\cite{P2P}, PnP~\cite{PnP}, MasaCtrl~\cite{MasaCtrl}, FlexiEdit~\cite{FlexiEdit}, and FreeDiff~\cite{FreeDiff}), FLUX-based approaches (RF-Inv~\cite{RF-Inv}, StableFlow~\cite{StableFlow}, RF-Edit~\cite{Rf-solver}, and DCEdit~\cite{DCEdit}), and DiT-based methods (FTEdit~\cite{FTEdit}, FlowEdit~\cite{FlowEdit}, and DNAEdit~\cite{DNAEdit}).
All models are evaluated using their publicly released implementations and default settings to ensure fair and consistent comparison.

\textbf{Metrics.}
To comprehensively evaluate both editing performance and background preservation, we adopt six complementary metrics. Structure Distance~\cite{StructureDistance} quantifies structural consistency between the edited and original images. PSNR, LPIPS~\cite{LPIPS}, MSE, and SSIM~\cite{SSIM} jointly assess content fidelity in unedited regions. For text-image alignment, we calculate CLIP similarity~\cite{CLIPsimilarity} over both the entire image and the specifically edited regions. Note that region masks provided by the dataset are used solely for evaluation purposes to isolate the edited areas.

\begin{table*}[htbp]
  \centering
  \small
  \begin{tblr}{
    colspec={c c c cccc cc c},
    vline{2,3,4,8,10} = {1-Z}{},
    hline{1,Z}={1pt},
    hline{2}={3-10}{},
    cell{1}{1}={r=2}{c},
    cell{1}{2}={r=2}{c},
    cell{1}{4}={c=4}{c},
    cell{1}{8}={c=2}{c} 
  }

    Method & Model &      Structure & Background Preservation & & & & CLIP Similarity & & Rank \\
    & &Distance$_{\times 10^3}\downarrow$ & PSNR $\uparrow$ & LPIPS$_{\times 10^3}\downarrow$ & MSE$_{\times 10^4}\downarrow$ & SSIM$_{\times 10^2}\uparrow$ & Whole $\uparrow$ & Edited $\uparrow$ & Avg. $\downarrow$ \\
    \hline
    P2P       & SD1.4 & $11.65$\textcolor{gray}{$^{\bm{2}}$} & $27.22$\textcolor{gray}{$^{\bm{2}}$} & $54.55$\textcolor{gray}{$^{\bm{1}}$} & $32.86$\textcolor{gray}{$^{\bm{3}}$} & $84.76$ & $25.02$ & $22.10$ & $5.3$ \\
    PnP       & SD1.5 & $24.29$ & $22.46$ & $106.06$ & $80.45$ & $79.68$ & $25.41$ & $22.62$ & $9.6$ \\
    MasaCtrl  & SD1.4 & $24.70$ & $22.64$ & $87.94$ & $81.09$ & $81.33$ & $24.38$ & $21.35$ & $10.7$ \\
    FlexiEdit & SD1.4 & $22.13$ & $25.74$ & $80.45$ & $58.45$ & $82.62$ & $25.15$ & $22.87$\textcolor{gray}{$^{\bm{2}}$} & $6.0$ \\
    FreeDiff  & SD1.5 & $18.70$ & $24.73$ & $89.76$ & $55.32$ & $81.68$ & $25.03$ & $22.12$ & $7.9$ \\
    \hline
    RF-Inv      & FLUX & $48.76$ & $19.51$ & $195.85$ & $155.74$ & $68.95$ & $25.11$ & $22.50$ & $11.6$ \\
    StableFlow  & FLUX & $19.24$ & $23.04$ & $76.94$ & $84.85$ & $87.22$ & $24.30$ & $21.28$ & $8.9$ \\
    RF-Edit     & FLUX & $27.70$ & $23.22$ & $131.18$ & $75.00$ & $81.44$ & $25.22$ & $22.40$ & $9.4$ \\
    DCEdit      & FLUX & $22.36$ & $25.41$ & $94.17$ & $48.09$ & $85.60$ & $25.47$ & $22.71$ & $6.1$ \\
    FTEdit      & SD3.5 & $18.17$ & $26.62$ & $80.55$ & $40.24$ & $91.50$\textcolor{gray}{$^{\bm{1}}$} & $25.74$\textcolor{gray}{$^{\bm{3}}$} & $22.27$ & $4.4$\textcolor{gray}{$^{\bm{3}}$} \\
    FlowEdit    & SD3.5 & $23.62$ & $23.21$ & $93.81$ & $69.95$ & $85.09$ & $26.78$\textcolor{gray}{$^{\bm{1}}$} & $23.73$\textcolor{gray}{$^{\bm{1}}$} & $6.1$ \\
    DNAEdit     & SD3.5 & $14.19$\textcolor{gray}{$^{\bm{3}}$} & $26.66$\textcolor{gray}{$^{\bm{3}}$} & $74.57$\textcolor{gray}{$^{\bm{3}}$} & $32.76$\textcolor{gray}{$^{\bm{2}}$} & $88.63$\textcolor{gray}{$^{\bm{3}}$} & $25.63$ & $22.71$ & $3.1$\textcolor{gray}{$^{\bm{2}}$} \\
    Ours        & SD3.5 & $10.34$\textcolor{gray}{$^{\bm{1}}$} & $27.32$\textcolor{gray}{$^{\bm{1}}$} & $55.02$\textcolor{gray}{$^{\bm{2}}$} & $28.66$\textcolor{gray}{$^{\bm{1}}$} & $89.21$\textcolor{gray}{$^{\bm{2}}$} & $25.89$\textcolor{gray}{$^{\bm{2}}$} & $22.82$\textcolor{gray}{$^{\bm{3}}$} & $1.7$\textcolor{gray}{$^{\bm{1}}$} \\
    
    \end{tblr}
    \caption{Quantitative comparison on PIE-Bench. Rank denotes the average ranking across all evaluation metrics. Our method achieves strong performance in both background preservation and semantic alignment, yielding the best average rank. Superscripts \textcolor{gray}{$^{\bm{1}}$}, \textcolor{gray}{$^{\bm{2}}$}, and \textcolor{gray}{$^{\bm{3}}$} denote the best, second-best, and third-best performance, respectively.}
    \label{tab:sota}
\end{table*}

\textbf{Implementation Details.}
FIA-Edit is based on the SD3.5-Medium~\cite{SD35} model with 50 sampling steps. Velocity fields are computed using CFG scales of 3.5 (source) and 13.5 (target). The FRI module is applied across all 50 steps, while the FIJ module is activated only during the first 27 steps to constrain early velocity field. 
During the $\mathbf{x}^{FE}_t$ stepping process, we add the reused interpolation noise $\epsilon_t$ with a scaling factor of $\sigma_t$.
Details of the Gaussian low-pass filter are provided in the Appendix. All experiments are run on a single NVIDIA RTX 4090 GPU. For comparison, we reproduce results of open-source baselines using their official code and default settings (see Appendix for details), while results of non-released methods are directly reported from their original papers.

\subsection{Comparisons with Other Editing Methods}
\textbf{Quantitative Comparison.} As shown in Table~\ref{tab:sota}, we conduct comprehensive evaluations on PIE-Bench across representative LDM-, FLUX-, and DiT-based methods. FIA-Edit achieves the best performance in background preservation while also demonstrating strong semantic alignment. Compared to the inversion-free baseline FlowEdit, our method preserves background details more accurately, highlighting the effectiveness of integrating source-target feature interactions during velocity field computation. Among inversion-based methods, P2P maintains relatively good background consistency but suffers from weak prompt alignment, suggesting that it overly retains source content. Overall, FIA-Edit delivers both superior background fidelity and precise semantic edits, leading to the best average ranking across all metrics, which validates the effectiveness of our approach.

\begin{table*}[htbp]
  \centering
  \small
  \begin{tblr}{
     colspec={c ccccc ccc cc},
    vline{2} = {1-Z}{},
    vline{7,10}={1-Z}{dashed},
    hline{1,Z}={1pt}
}
    
    Method &   P2P &   PnP &   MasaCtrl    &   FlexiEdit   &   FreeDiff&
    \textit{RF-Inv}  &   \textit{StableFlow}  &   \textit{RF-Edit} &   FlowEdit    &   Ours\\
    \hline
    GPU(GB) &$10.95$& $8.99$&   $11.42$     &   $18.73$     &   $6.08$  &
            $\textit{69.22}$ & $\textit{35.39}$ & $\textit{32.91}$ &   $17.93$     &   $17.93$\\
    Time(s) &$34.84$&$18.09$&   $21.71$     &   $38.97$     &   $17.41$ &
            $\textit{76.74}$ & $\textit{26.07}$ & $\textit{34.51}$  &   $3.49$      &   $6.30$\\
    
  \end{tblr}
  \caption{Memory and runtime comparison. RF-Inv, StableFlow, and RF-Edit were run on an A100 80GB GPU, while all other methods were tested on a single RTX 4090. Our approach achieves a favorable balance between speed and editing quality.}
  \label{tab:runtime}
\end{table*}

\begin{table*}[htbp]
  \centering
  \small
  \begin{tblr}{
  colspec={cc|c|cccc|cc},
  cells = {halign=c, valign=m},
  hline{1,Z}={1pt},
  cell{1}{1}={c=2}{c},
  cell{1}{4}={c=4}{c},
  cell{1}{8}={c=2}{c}
  }
    Module& &   Structure                         
    & Background Preservation &  & & & CLIP Similarity& \\
    \hline
    FIJ  &   FRI    &  Distance$_{\times10^3}\downarrow$  
    &PSNR $\uparrow$
    &LPIPS$_{\times10^3}\downarrow$
    &MSE$_{\times10^4}\downarrow$
    &SSIM$_{\times10^2}\uparrow$
    &Whole $\uparrow$&Edited $\uparrow$\\
    \hline
    $\times$ &   $\times$ &    $23.62$ 
    &   $23.21$ &   $93.81$ &   $69.95$ &   $85.09$
    &   $26.78$ &   $23.73$\\
    $\checkmark$ &   $\times$ &    $14.89$ 
    &   $25.59$ &   $70.18$ &   $41.74$ &   $87.51$
    &   $26.30$ &   $23.12$\\    
    $\checkmark$ &   $add$ &    $16.50$ 
    &   $25.93$ &   $85.44$ &   $38.72$ &   $86.51$
    &   $26.05$ &   $22.68$\\
    $\checkmark$ &   $freq$ &    $10.34$ 
    &   $27.32$ &   $55.02$ &   $28.66$ &   $89.21$
    &   $25.89$ &   $22.82$\\

  \end{tblr}
  \caption{Ablation study on key components of FIA-Edit. FRI and FIJ denote the proposed Frequency Representation Interaction and Feature Injection modules, respectively. Within FRI, $freq$ refers to our frequency-domain fusion design, while $add$ denotes direct addition of source and target features.}
  \label{tab:ablation}
\end{table*}

\begin{table*}[htbp]
  \centering
  \small
  \begin{tblr}{
  colspec={c c c cccc cc},
  vline{2,3,4,8}={1-Z}{dashed},
  vline{2}={1-Z}{},
  hline{1,Z}={1pt}
  }

  Method   &   ConvNeXt-T  & Aug &  PnP &   MasaCtrl    &   FlexiEdit   &   FreeDiff    
  &   FlowEdit    &   Ours\\
  \hline
  AUC (\%) & $81.54$ & $82.10$ & $81.98$ & $\underline{84.22}$ & $82.05$ & $82.25$ & $83.83$ & $\mathbf{85.05}$\\
  PR-AUC (\%) & $38.66$ & $38.81$ & $38.53$ & $38.82$ & $37.97$ & $38.60$ & $\underline{40.34}$ & $\mathbf{43.81}$\\
  Precision (\%) & $50.90$ & $49.18$ & $53.03$ & $51.92$ & $52.62$ & $\underline{53.68}$ & $50.88$ & $\mathbf{54.01}$\\
  Recall (\%) & $29.49$ & $30.84$ & $26.57$ & $27.17$ & $26.03$ & $25.65$ & $\underline{31.44}$ & $\mathbf{32.90}$\\
  F1-score (\%) & $37.35$ & $37.91$ & $35.40$ & $35.67$ & $34.83$ & $34.71$ & $\underline{38.86}$ & $\mathbf{40.89}$\\
  Accuracy (\%) & $91.93$ & $91.76$ & $92.09$ & $92.01$ & $92.05$ & $\underline{92.13}$ & $91.93$ & $\mathbf{92.24}$\\
  \end{tblr}
  \caption{Comparison of bleeding classification performance. All methods except ConvNeXt-T augment bleeding data with an additional $\sim$5,000 images. Aug denotes traditional augmentation. Ours significantly improves Recall, showing the value of editing-based augmentation. \textbf{Bold}: best; \underline{underline}: second-best.}
  \label{tab:downstream}
\end{table*}

\begin{figure*}
\begin{center}
	\includegraphics[width=0.835\linewidth]{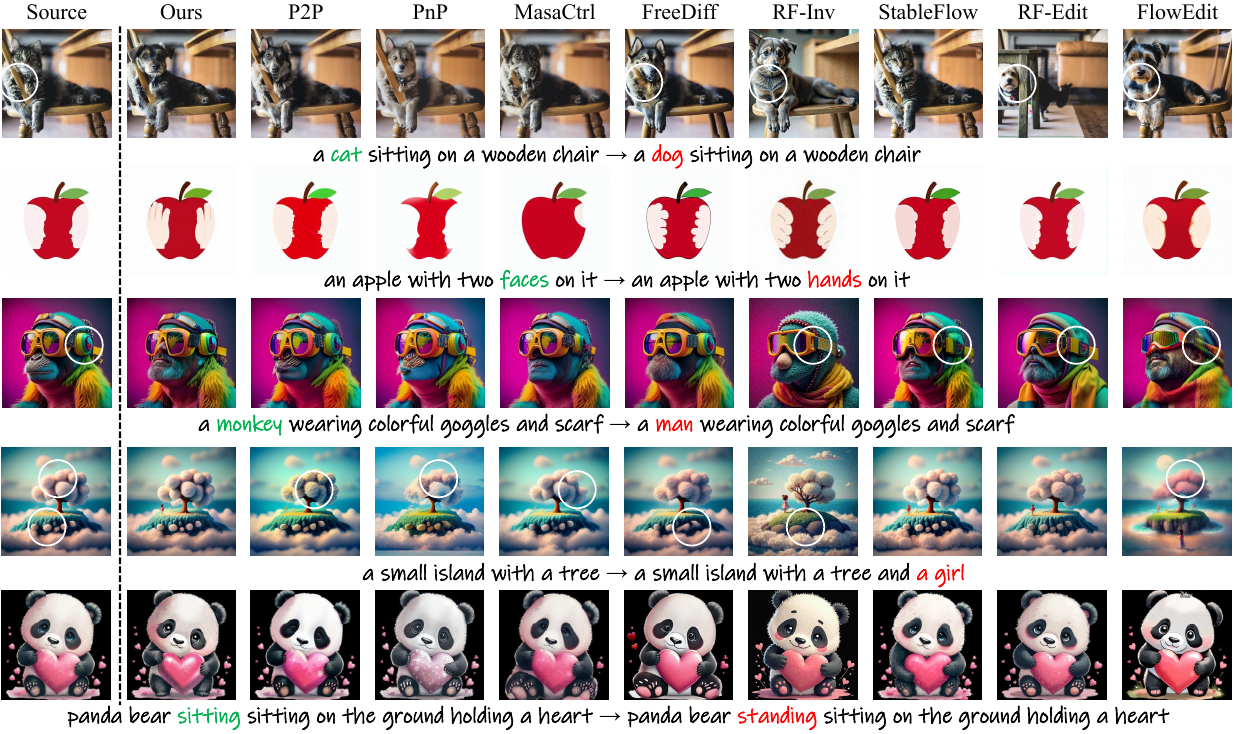}
\end{center}
\caption{
Qualitative comparison. Our method preserves the background while accurately reflecting the target semantics. White circles highlight cases where other methods poorly preserve non-editing regions.
}
\label{fig:Vis}
\end{figure*}

\textbf{Qualitative Comparison.} Visual results are shown in Fig.~\ref{fig:Vis}, covering content alterations, object addition, and pose change. Our method produces high-quality, semantically accurate edits while maintaining background integrity. In contrast, other methods either fail to achieve the intended semantic change or suffer from noticeable background distortion (\textit{e.g.}, FlowEdit), clearly demonstrating the superiority of our approach.

\textbf{GPU Memory and Runtime.} 
We report GPU memory and runtime of open-source methods in Table~\ref{tab:runtime}, measured on a single RTX 4090 for fair comparison. Due to high memory demands, RF-Inv, StableFlow, and RF-Edit are tested on an A100 (80GB). Runtime is averaged over 10 samples (image size: $512 \times 512$), covering the full pipeline from loading to saving. As shown, inversion-free methods are notably faster. Compared to FlowEdit, our method incurs slightly more runtime due to feature interaction, while keeping memory usage comparable. Overall, FIA-Edit strikes a strong balance between quality and efficiency.

\subsection{Ablation Study}
We conduct ablations to evaluate the effectiveness of Frequency Representation Interaction (FRI) and Feature Injection (FIJ), as shown in Table~\ref{tab:ablation}. In FRI, $freq$ denotes our frequency-domain fusion, while $add$ refers to simple feature addition.
Row 1 vs. Row 2 shows that FIJ greatly improves background preservation by injecting source features during velocity estimation.
Row 2 vs. Row 4 indicates that combining FRI on top of FIJ further enhances semantic alignment while retaining strong background consistency, demonstrating the value of frequency-guided interaction.
Row 3 vs. Row 4 shows that our structured $freq$ design outperforms naive $add$.
Note that CLIP score alone does not reflect editing quality, ideal edits require both high semantic fidelity and accurate background preservation. See Appendix for visual examples illustrating this balance.

\section{Editing for Bleeding Classification Task}

\textbf{Clinical Motivation.}
Early detection of abnormal intraoperative bleeding is crucial yet challenging. Identifying early bleeding from surgical videos can assist surgeons in rapidly locating bleeding sites. However, such cases are rare, leading to severe data imbalance. Existing efforts on surgical image or video synthesis rarely address bleeding scenarios, let alone via image editing. To the best of our knowledge, we are the first to explore \textit{text-guided image editing} for surgical bleeding augmentation. Our approach aims to enrich bleeding variations and mitigate data imbalance, ultimately improving downstream classification performance.

\textbf{Experimental Setup.} 
We use the Laparoscopic Roux-en-Y Gastric Bypass dataset~\cite{BloodDataset} with 140 videos (100 for training, 40 for testing), sampled at 1fps for 770K frames. The training set includes 512K normal and 44K bleeding frames; the test set has 197K normal and 17K bleeding frames, indicating severe imbalance.
We adopt ConvNeXt-T~\cite{ConvNeXt} as the classification backbone. From the training set, we extract 4,803 early-stage bleeding frames ($\sim$50 per video) and edit them into varying bleeding levels (see Appendix). To ensure editing quality and efficiency, we compare four LDM-based and two inversion-free methods, plus standard augmentations (\textit{e.g.}, flipping, rotation). All editing models use SD 1.5 or 3.5 checkpoints without task-specific finetuning. The augmented images are incorporated into training set for downstream classification.

\textbf{Results and Analysis.} Quantitative results are shown in Table 4, with visualizations provided in the Appendix. Traditional augmentation methods yield only marginal gains despite generating an additional 5K frames. The four inversion-based methods improve precision but degrade recall, indicating reduced sensitivity to bleeding. This may stem from limited generation quality of SD1.5, potentially requiring domain-specific finetuning. In contrast, FlowEdit and our method both improve recall, suggesting better utility in enhancing bleeding classification. However, FlowEdit’s poor background preservation may lead to false positives. Benefiting from both semantic fidelity and structural consistency, our method achieves the most balanced performance, demonstrating the potential of image editing techniques for surgical data augmentation and downstream applications.




\section{Conclusion}

We present FIA-Edit, a novel and efficient inversion-free image editing framework that introduces Frequency-Interactive Attention for improved semantic alignment and background preservation. By explicitly modeling source-target interactions in both the frequency and spatial domains, our approach addresses key limitations of existing inversion-free methods, achieving high-quality, controllable image edits without costly inversion.
To the best of our knowledge, we are the first to explore the use of generative image editing for \textit{clinical data augmentation}. Specifically, we apply FIA-Edit to synthesize plausible variations of surgical bleeding images, resulting in improved performance on downstream classification task. This demonstrates the potential of controlled image editing in medical data scenarios, opening new avenues for future research.
We hope our work contributes a new perspective to tuning-free editing and inspires broader exploration of generative techniques in real-world applications.




\clearpage
\section{Acknowledgments}
This work was supported in part by National Key R\&D Program of China (Grant No. 2023YFC2414900), National Natural Science Foundation of China (Grant No.62202189), and research grants from Wuhan United Imaging Healthcare Surgical Technology Co., Ltd.
\bibliography{aaai2026}

\clearpage

\appendix




\section{Preliminary}

In this section, we briefly introduce the two key components that form the foundation of our tuning-free FIA-Edit framework: Rectified Flow~\cite{Flow1,Flow2,Flow3} and Classifier-Free Guidance~\cite{CFG}.

\subsection{Rectified Flow}
Rectified flow~\cite{Flow1, Flow2, Flow3} aims to model a probability path between two distributions, $\mathbf{x}_0 \sim p_0$ and $\mathbf{x}_1 \sim p_1$, through linear interpolation. The continuous probability flow is defined as:
\begin{equation}
    \mathbf{x}(t) = (1 - t)\mathbf{x}_0 + t\mathbf{x}_1,\quad t \in [0, 1].
\end{equation}
To learn a straight probability transport path, Rectified flow formulates an ordinary differential equation (ODE), governed by a learnable velocity field $v_\theta(\mathbf{x}_t, t)$:
\begin{equation}
    \text{d}\mathbf{x}_t = {v}_\theta(\mathbf{x}_t, t)\text{d}t.
\end{equation}
In the standard setting, the source distribution $p_0$ is chosen from standard Gaussian $\mathcal{N}(\mathbf{0}, \mathbf{I})$. Sampling is performed by integrating the learned velocity field from $t = 0$ to $t = 1$, starting from $\mathbf{x}_0 \sim p_0$, to produce a sample $\mathbf{x}_1$.
In practice, the ODE is discretized via the Euler method, and the sample is updated iteratively as:
\begin{equation}
    \mathbf{x}_{t+1} = \mathbf{x}_{t} + (\sigma_{t+1} - \sigma_t){v}_\theta(\mathbf{x}_{t}, t),
\label{eq:RF-generate}
\end{equation}
where $\sigma_t$ denotes the discrete timestep.
This formulation ensures that the sampling trajectory remains approximately linear, contributing to improved stability and efficiency during the generation process.

\subsection{Classifier-Free Guidance}
To enhance conditional control in generative diffusion models, Ho \textit{et al.}~\cite{CFG} introduced Classifier-Free Guidance (CFG). Instead of relying on an external classifier, CFG achieves guidance by linearly interpolating between conditional and unconditional noise predictions.
Specifically, given a condition $c$, the model predicts noise as $\epsilon_\theta(x_t, c)$ when conditioned, and $\epsilon_\theta(x_t, \varnothing)$ when unconditioned.
CFG combines these predictions into a guided estimate:
\begin{equation} \label{CFG}
\tilde \epsilon_\theta(x_t,c) \;=\; \epsilon_\theta(x_t, \varnothing)\;+\;\mu \bigl(\epsilon_\theta(x_t,c)\;-\;\epsilon_\theta(x_t,\varnothing)\bigr),
\end{equation}
where $\mu>1$ controls the guidance strength.

\section{Algorithm of FIA-Edit}

Here, we present the algorithm of our proposed FIA-Edit. As shown below, our FIA Constraint enables cross-domain interaction between source and target features during the computation of their respective velocity fields.

\begin{algorithm}[ht]
\caption{Algorithm for FIA-Edit}
\label{alg:algorithm}
\textbf{Input}: source image $\mathbf{X}^{src}$, source prompt $\mathcal{P}^{src}$, target prompt $\mathcal{P}^{tar}$, editing steps $T$ \\
\textbf{Output}: edited target image $\mathbf{X}^{tar}$
\begin{algorithmic}[1] 
\STATE Let $\mathbf{x}^{FE}_T = \mathbf{X}^{src}$
\FOR{$t=T$ \textbf{to} $1$}
\STATE $\epsilon_t \sim \mathcal{N}(\mathbf{0},\mathbf{I})$ \hfill \textcolor{gray}{// Sample random Gaussian noise}
\STATE $\mathbf{x}^{src}_t = (1 - \sigma_t)\cdot \mathbf{X}^{src} + \sigma_t\cdot \epsilon_t$
\STATE $\mathbf{x}^{tar}_t = \mathbf{x}^{FE}_t + \mathbf{x}^{src}_t - \mathbf{X}^{src}$
\STATE $\{f^{src}\} \leftarrow v_\theta(\mathbf{x}^{src}_t, \mathcal{P}^{src}, t)$ \hfill \textcolor{gray}{// Intermediate features}
\STATE $\{f^{tar}\} \leftarrow v_\theta(\mathbf{x}^{tar}_t, \mathcal{P}^{tar}, t)$ \hfill \textcolor{gray}{// Intermediate features}

\STATE $v^{\Delta}_t = v_\theta(\mathbf{x}^{tar}_t, \mathcal{P}^{tar}, t, \mathtt{FIA}(\{f^{src}\},\{f^{tar}\})) - v_\theta(\mathbf{x}^{src}_t, \mathcal{P}^{src}, t)$ \hfill \textcolor{gray}{// Constrain the target velocity field}
\STATE $\mathbf{x}^{FE}_{t-1} = \mathbf{x}^{FE}_t + (\sigma_{t-1} - \sigma_t)\cdot v^{\Delta}_t + \sigma_t\cdot \epsilon_t$
\ENDFOR
\STATE \textbf{Return} $\mathbf{X}^{tar} = \mathbf{x}^{FE}_0$
\end{algorithmic}
\end{algorithm}

\section{Comparison Methods and Details}
In this section, we detail the experimental setup and hyperparameter configurations for the comparison methods used in the main paper.

\subsection{LDM-Based Methods}

For P2P~\cite{P2P}, PnP~\cite{PnP}, and MasaCtrl~\cite{MasaCtrl}, we adopt DDIM Direct Inversion~\cite{PIE-Bench-DirInvPmt}\footnote{\url{https://github.com/cure-lab/PnPInversion}} as the inversion backbone. All editing experiments are conducted using their default parameter settings, following the above official implementation.

For FlexiEdit~\cite{FlexiEdit}, we use the official codebase\footnote{\url{https://github.com/kookie12/FlexiEdit}} and fix the reinversion steps to $t_R = 30$ for stable and consistent batch editing, as suggested by the most common official usage. The required `blended word' for localized editing is extracted from corresponding prompts in PIE-Bench~\cite{PIE-Bench-DirInvPmt}. If the model fails to identify the relevant semantic region, we default to full-image editing. Other settings remain unchanged. 

For FreeDiff~\cite{FreeDiff}, we follow the official implementation\footnote{\url{https://github.com/Thermal-Dynamics/FreeDiff}} with the recommended configuration: filter scheduling to $\tau_i = (801, 781, 581)$, and the high-pass filter sizes to $r_t^H = (32, 32, 10, 10)$. All other parameters use default values.

All LDM-based methods above are based on Stable Diffusion v1.4 or v1.5 and executed on a single NVIDIA RTX 4090 GPU with 24GB of memory.

\subsection{DiT-Based Methods}

For StableFlow~\cite{StableFlow}, we adopt the official implementation\footnote{\url{https://github.com/snap-research/stable-flow}} with inversion steps set to 50.

For RF-Inv~\cite{RF-Inv}, we use the official codebase\footnote{\url{https://github.com/LituRout/RF-Inversion}} and retain all default settings.

For RF-Edit~\cite{Rf-solver}, we follow the official repository\footnote{ \url{https://github.com/wangjiangshan0725/RF-Solver-Edit}}, setting the guidance scale to 2 and the injection step to 5. All other parameters are left unchanged. The above three methods are executed on a single NVIDIA A100-PCIE-80GB GPU due to their higher memory requirements.

For FlowEdit~\cite{FlowEdit}, we use the official codebase\footnote{ \url{https://github.com/fallenshock/FlowEdit}} with all default settings and run all experiments on a single NVIDIA RTX 4090 GPU with 24GB of memory.

\section{User Study}
To evaluate editing quality from a subjective perspective, we conducted a user study. We selected about 20 images along with the edited results produced by P2P~\cite{P2P}, FreeDiff~\cite{FreeDiff}, StableFlow~\cite{StableFlow}, FlowEdit~\cite{FlowEdit}, and Ours, and designed a questionnaire in which participants ranked the five methods for each image. A higher score indicates better perceived editing quality, with a maximum of 5 points.

We collected 16 valid responses, and the aggregated results are shown in Table~\ref{tab:UserStudy}. The results clearly indicate that our method achieves the highest subjective scores, aligning well with both the quantitative metrics and qualitative visual comparisons presented in the paper.

\begin{table}
  \centering
  \begin{tabular}{c c}
  \toprule
  Method & Score \\
  \midrule
  P2P         & $2.19 \pm 0.86$ \\
  FreeDiff    & $2.21 \pm 0.74$ \\
  StableFlow  & $2.71 \pm 0.69$ \\
  FlowEdit    & $3.14 \pm 0.79$ \\
  Ours        & $\mathbf{4.45 \pm 0.38}$ \\
  \bottomrule
  \end{tabular}
  \caption{User study results. Higher scores indicate better perceived editing quality.}
  \label{tab:UserStudy}
\end{table}  

\section{More Ablation Studies}

\subsection{Visualization of ablation study for FIA}

\begin{figure*}[htbp]
\begin{center}
	\includegraphics[width=1\linewidth]{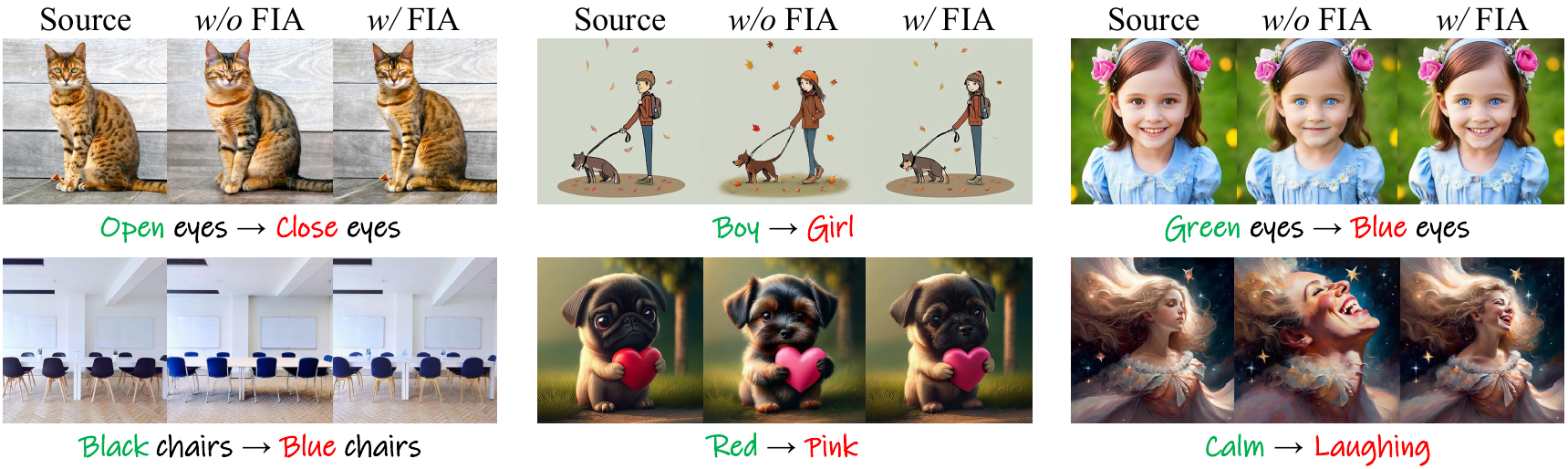}
\end{center}
\caption{
Qualitative comparison of editing results with and without the proposed FIA constraint. Each unit shows a source image (left), an edited result without FIA (middle, $w/o$ FIA), and an edited result with FIA (right, $w/$ FIA). The FIA constraint significantly improves background preservation and enhances semantic accuracy in the edited regions.
}
\label{fig:SM1-aba}
\end{figure*}

To better illustrate the impact of our FIA constraint on background preservation within inversion-free editing, we provide qualitative comparisons with and without the constraint in Fig.~\ref{fig:SM1-aba}. Each of the six units visualizes one example, with the three images from left to right showing the source image, editing without FIA ($w/o$ FIA), and editing with our proposed constraint ($w/$ FIA), respectively. As shown, applying FIA significantly improves background preservation in non-edited regions. Moreover, in the bottom-right example, the edited result with FIA also aligns better with the target semantics. Together with the quantitative results in the main paper, this qualitative comparison further highlights the effectiveness of our FIA constraint in achieving semantically accurate and spatially coherent edits in inversion-free pipelines.

\subsection{FIJ Placement Strategy}
As described in the main paper, the FRI module is used in all self-attention layers. Here, we further investigate where to place the FIJ module within the DiT architecture. The SD 3.5 Medium variant contains 24 Transformer blocks in total: blocks $0\sim12$ include both self- and cross-attention, while blocks $13\sim 23$ contain only cross-attention.

We explore three configurations for injecting FIJ: early blocks ($0\sim 12$), late blocks ($13\sim 23$), and all blocks ($0\sim 23$). As shown in Table~\ref{tab:FIJ_Replace}, employing FIJ only in early blocks fails to preserve background well, likely due to limited semantic abstraction in shallow features. In contrast, applying FIJ across all blocks results in overly strong background preservation, suppressing necessary semantic changes. The best trade-off is achieved by applying FIJ into the deeper cross-attention blocks $13\sim  23$, where high-level features guide editing.

\begin{table*}
  \centering
  \small
  \begin{tblr}{
    colspec={c c cccc cc c},
    vline{2,3,7} = {1-Z}{},
    hline{1,Z}={1pt},
    hline{2}={2-9}{},
    cell{1}{1}={r=2}{c},
    cell{1}{3}={c=4}{c},
    cell{1}{7}={c=2}{c} 
  }

    FIJ locations &  Structure & Background Preservation & & & & CLIP Similarity & \\
    & Distance$_{\times 10^3}\downarrow$ & PSNR $\uparrow$ & LPIPS$_{\times 10^3}\downarrow$ & MSE$_{\times 10^4}\downarrow$ & SSIM$_{\times 10^2}\uparrow$ & Whole $\uparrow$ & Edited $\uparrow$ \\
    \hline
    $0\sim12$  & $58.66$ & $18.87$ & $200.51$ & $173.71$ & $75.12$ & $25.69$ & $22.09$ \\   
    $13\sim  23$ (Ours) &  ${10.34}$ & ${27.32}$ & ${55.02}$ & ${28.66}$ & ${89.21}$ & ${25.89}$ & $22.82$ \\
    $0\sim23$   & $10.11$ & $27.39$ & $54.49$ & $28.33$ & $89.18$ & $25.78$ & $22.72$ \\
    \end{tblr}
    \caption{Performance under FIJ locations. $0\sim 12$, $13 \sim 23$, and $0 \sim 23$ indicate the index ranges of DiT transformer blocks where FIJ is injected.}
    \label{tab:FIJ_Replace}
\end{table*}

\subsection{Noise Reuse During Stepping}
To further investigate the role of noise during the intermediate feature stepping process, we compare three strategies: (1) no noise added ($w/o~\mathcal{N}(\mathbf{0},\mathbf{I})$), (2) random Gaussian noise at each step ($\sigma_t \cdot \mathcal{N}(\mathbf{0},\mathbf{I})$), and (3) reusing the noise from the initial sampling step ($\sigma_t \cdot \epsilon_t$). Results are shown in Table~\ref{tab:NoiseStep}.

We observe that injecting random noise during the feature update process degrades background fidelity, likely due to the additional stochasticity that induces source-agnostic synthesis. In contrast, reusing the initial noise already aligned with the source image effectively preserves background consistency. This reused noise implicitly encodes the velocity information of background elements during interpolation from source to target. Our approach strikes a better balance between semantic change and background retention.

\begin{table*}
  \centering
  \small
  \begin{tblr}{
    colspec={c c cccc cc c},
    vline{2,3,7} = {1-Z}{},
    hline{1,Z}={1pt},
    hline{2}={2-9}{},
    cell{1}{1}={r=2}{c},
    cell{1}{3}={c=4}{c},
    cell{1}{7}={c=2}{c} 
  }

    Noise Stepping &  Structure & Background Preservation & & & & CLIP Similarity & \\
    & Distance$_{\times 10^3}\downarrow$ & PSNR $\uparrow$ & LPIPS$_{\times 10^3}\downarrow$ & MSE$_{\times 10^4}\downarrow$ & SSIM$_{\times 10^2}\uparrow$ & Whole $\uparrow$ & Edited $\uparrow$ \\
    \hline
    $w/o~\mathcal{N}(\mathbf{0},\mathbf{I})$  & $11.02$ & $27.17$ & $57.39$ & $29.95$ & $89.00$ & $25.92$ & $22.83$ \\   
    
    $\sigma_t \cdot \mathcal{N}(\mathbf{0},\mathbf{I})$   & $11.03$ & $27.03$ & $58.67$ & $30.56$ & $88.83$ & $25.95$ & $22.80$ \\

    $\sigma_t \cdot \epsilon_t$ (Ours) &  ${10.34}$ & ${27.32}$ & ${55.02}$ & ${28.66}$ & ${89.21}$ & ${25.89}$ & $22.82$ \\
    \end{tblr}
    \caption{Effect of Noise Stepping. $w/o~\mathcal{N}(\mathbf{0},\mathbf{I})$ disables noise injection during stepping, $\sigma_t \cdot \mathcal{N}(\mathbf{0},\mathbf{I})$ refers to newly sampled Gaussian noise at stepping, $\sigma_t \cdot \epsilon_t$ means reusing the initial noise from the sampling process.}
    \label{tab:NoiseStep}
\end{table*}

\subsection{Effect of Gaussian Filter Strength in FRI}

We employ a Gaussian low-pass filter $\mathcal{L}$ with a scaling coefficient $\sigma$ in the FRI module, defined as:
\begin{equation}
    \mathcal{L}_\sigma = \frac{1}{2\pi\sigma^2} e^{-\frac{r^2}{2\sigma^2}} \in \mathbb{R}^{W \times H},
\end{equation}
where $\sigma$ controls the degree of Gaussian curve.

To evaluate its influence on the overall editing performance, we conduct an ablation study by varying $\sigma \in \{0.2, 0.4, 0.8, 0.9, 1.0, 1.5, 5.0, 10.0\}$, as summarized in Table~\ref{tab:Sigma}. Results show that as $\sigma$ increases, both background preservation and semantic fidelity initially improve, then degrade. We select $\sigma = 0.9$ as the optimal setting, where FIA achieves the best trade-off by effectively constraining the target velocity field using informative cues from the source velocity field.

\begin{table*}[htbp]
  \centering
  \small
  \captionsetup{singlelinecheck=off, justification=raggedright}
  \begin{tblr}{
  colspec={c|c|cccc|cc},
  hline{1,Z}={1pt},
  hline{2}={2-8}{},
  cell{1}{1}={r=2}{c},
  cell{1}{3}={c=4}{c},
  cell{1}{7}={c=2}{c}
  }

    $\sigma$  &   Structure                         
    &   Background Preservation& & & &   CLIP Similarity& \\

     &    Distance$_{\times10^3}\downarrow$  
    &PSNR $\uparrow$
    &LPIPS$_{\times10^3}\downarrow$
    &MSE$_{\times10^4}\downarrow$
    &SSIM$_{\times10^2}\uparrow$
    &Whole $\uparrow$&Edited $\uparrow$\\
    \hline
    $0.2$       &   $10.87$  &   $27.16$ &   $57.13$ &   $29.96$
    &   $88.97$ &   $25.85$  &   $22.74$\\
    $0.4$       &   $10.86$  &   $27.16$ &   $57.12$ &   $29.93$
    &   $88.98$ &   $25.86$  &   $22.74$\\
    $0.8$     &   $10.59$  &   $27.24$ &   $55.99$ &   $29.33$
    &   $89.09$ &   $25.89$ &   $22.76$\\
    $0.9$ (Ours) & $10.34$ & $27.32$ & $55.02$ & $28.66$ & $89.21$ & $25.89$ & $22.82$ \\
    $1.0$     &   $10.50$ &   $27.27$ &   $55.55$ &   $29.06$
    &   $89.14$ &   $25.88$ &   $22.75$\\
    $1.5$     &   $10.64$ &   $27.23$ &   $56.16$ &   $29.42$
    &   $89.08$ &   $25.87$ &   $22.78$\\
    $5.0$     &   $10.85$ &   $27.16$ &   $57.05$ &   $29.90$   
    &   $88.98$ &   $25.85$ &   $22.71$\\
    $10.0$     &   $10.83$ &   $27.12$ &   $57.59$ &   $30.01$
    &   $88.95$ &   $25.83$ &   $22.75$\\
  \end{tblr}
  \caption{Impact of Gaussian filter parameter $\sigma$ on editing performance. We vary the low-pass filter scale $\sigma$ in the FRI module to study its effect on background preservation and semantic consistency. A moderate value ($\sigma = 0.9$) achieves the best trade-off, effectively guiding the target velocity field using source frequency cues.}
  \label{tab:Sigma}
\end{table*}

\begin{figure*}[htbp]
\begin{center}
	\includegraphics[width=1\linewidth]{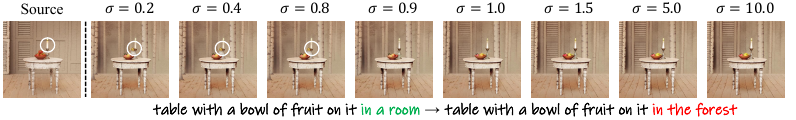}
\end{center}
\caption{
Effect of different $\sigma$ values on editing results. This figure illustrates how varying the Gaussian low-pass filter parameter $\sigma$ influences the output. Small $\sigma$ values (\textit{e.g.}, 0.2, 0.4) fail to preserve background structures like the candlestick, while large values (\textit{e.g.}, 1.5, 5.0) hinder effective replacement with the target semantic (\textit{e.g.}, forest background). A value of $\sigma = 0.9$ achieves the best trade-off between background preservation and semantic alignment.
}
\label{fig:SM2-sigma}
\end{figure*}

We further visualize the impact of different $\sigma$ values in the Gaussian low-pass filter on editing results, as shown in Fig.~\ref{fig:SM2-sigma}. As $\sigma$ increases, the editing quality first improves and then degrades. Specifically, when $\sigma < 0.9$, key background structures such as the candlestick are poorly preserved, indicating insufficient guidance from the source features. Conversely, when $\sigma > 0.9$, the model struggles to apply semantic changes, for instance, it fails to replace the background with a forest scene, suggesting that the target features are overly suppressed. A value of $\sigma = 0.9$ achieves the best trade-off between background fidelity and semantic consistency.

\section{Additional Visualization Results}

\subsection{Comparison with SOTA Methods}
\begin{figure*}[htbp]
\begin{center}
	\includegraphics[width=0.9\linewidth]{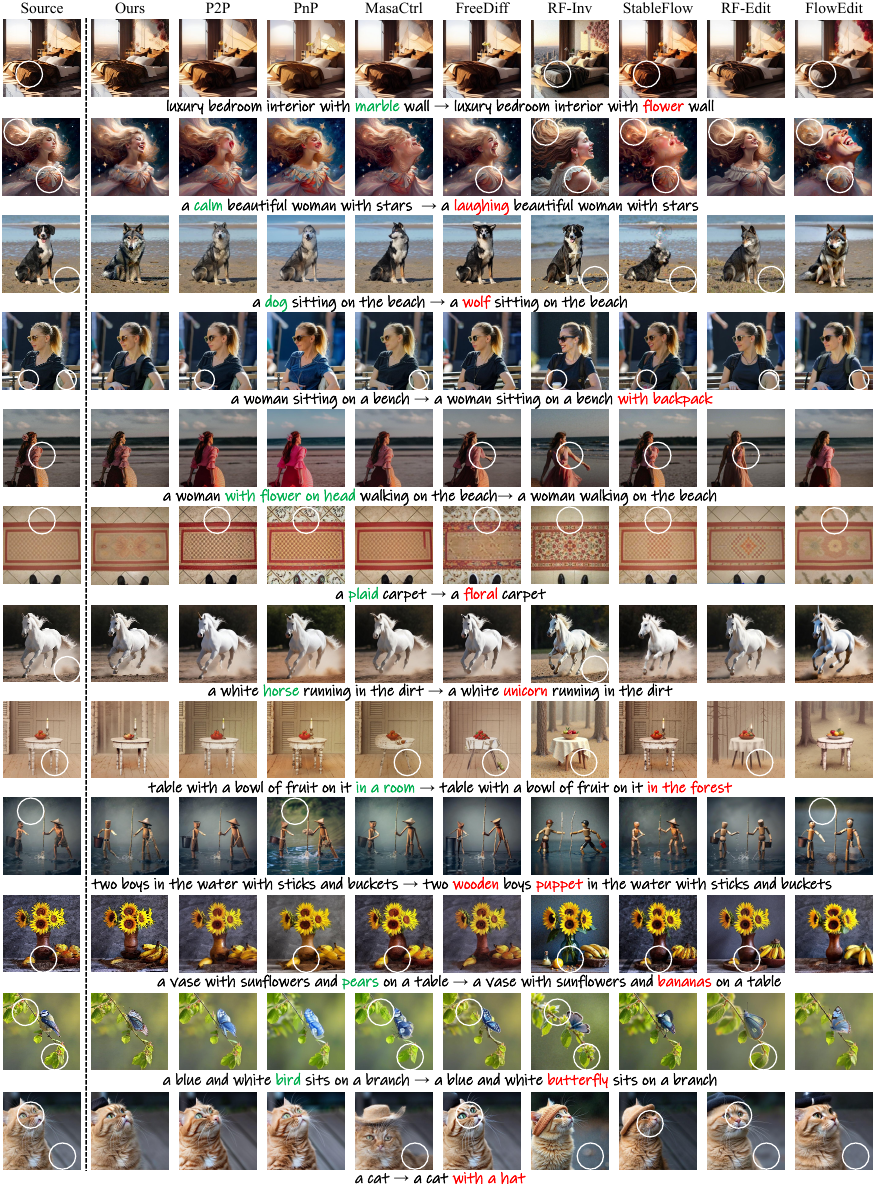}
\end{center}
\caption{
Additional comparisons with SOTA methods on PIE-Bench. White circles highlight cases where other methods poorly preserve non-editing regions, while visually unsatisfactory edits without such issues are not marked, demonstrating the superior background preservation of our method. (Zoom in for details.)
}
\label{fig:SM3-SOTAcomparison}
\end{figure*}

We present additional qualitative comparisons with existing methods in Fig.~\ref{fig:SM3-SOTAcomparison}. The editing scenarios cover a wide range of categories, including background replacement, object modification, addition and removal, as well as texture changes. White circles in the figure highlight areas where competing methods fail to preserve non-edited regions.
From the visual comparisons, we observe that several methods fail to perform semantically correct edits (\textit{e.g.}, P2P in the second row, or most methods in the eighth row). Others exhibit poor background preservation, with significant artifacts in non-edited areas, as marked by the white circles.
In particular, when compared to the FlowEdit backbone, our method demonstrates significantly better semantic alignment and background consistency, highlighting the effectiveness of our proposed approach.

\subsection{More Editing Types Visualization}
\begin{figure*}[htbp]
\begin{center}
	\includegraphics[width=1\linewidth]{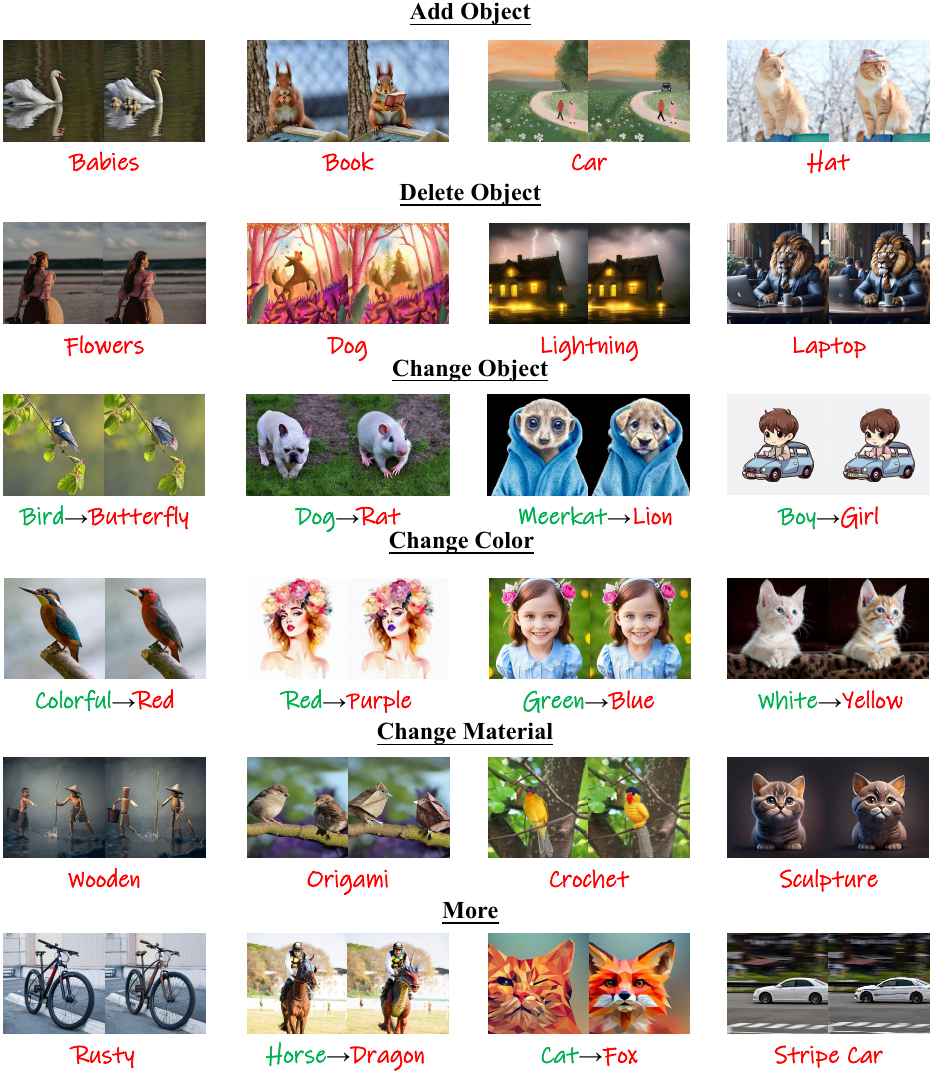}
\end{center}
\caption{
Additional qualitative results on diverse editing types from PIE-Bench Part I. (Zoom in for details.)
}
\label{fig:SM4-OursMore}
\end{figure*}

\begin{figure*}[htbp]
\begin{center}
	\includegraphics[width=1\linewidth]{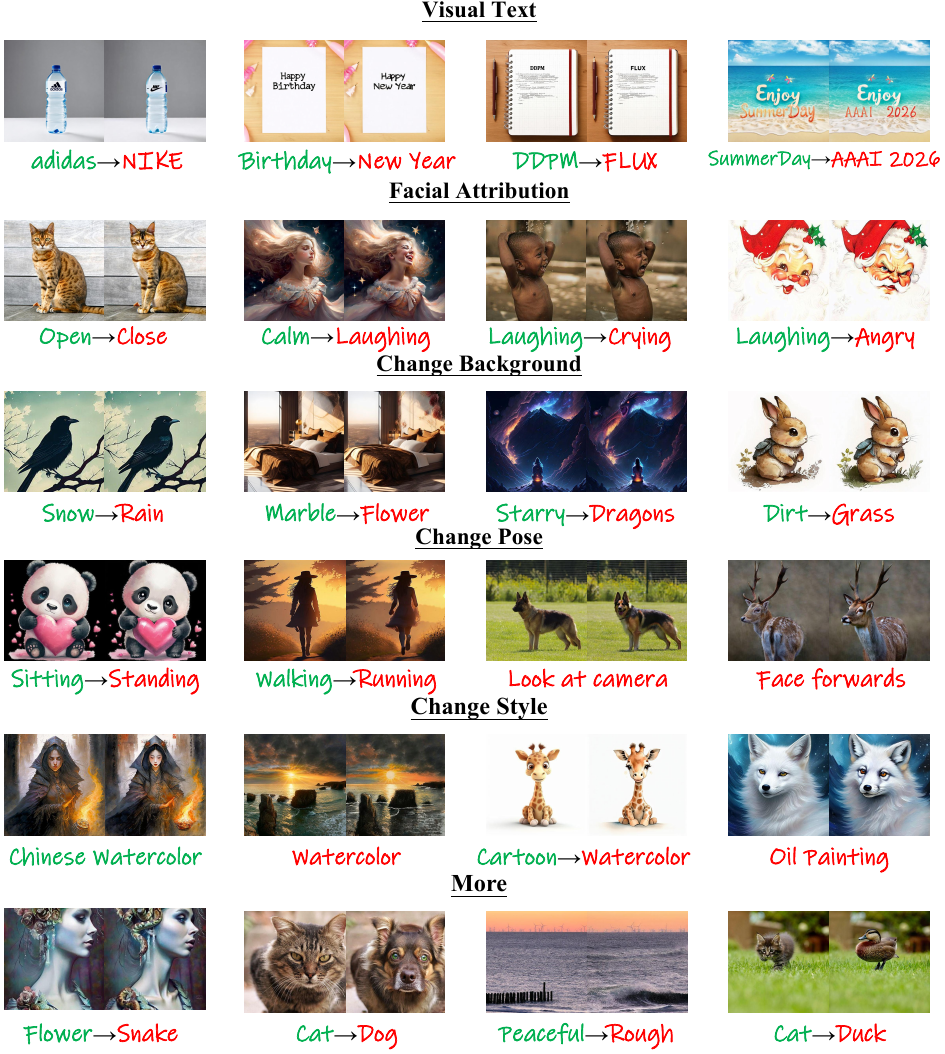}
\end{center}
\caption{
Additional qualitative results on diverse editing types from PIE-Bench Part II. (Zoom in for details.)
}
\label{fig:SM5-OursMore}
\end{figure*}

In Fig.~\ref{fig:SM4-OursMore} and Fig.~\ref{fig:SM5-OursMore}, we showcase a wider range of editing tasks performed by our method on the PIE-Bench dataset. These examples cover diverse categories, including text changes, facial expressions, object addition and removal, pose changes, style and color shifts, and material transformations.
Our method consistently produces edits that accurately follow the given semantic descriptions while effectively preserving non-edited background regions. This demonstrates the robustness and practical applicability of our approach across various scenarios.

\section{Failure Case Analysis}
\begin{figure*}[htbp]
\begin{center}
	\includegraphics[width=1\linewidth]{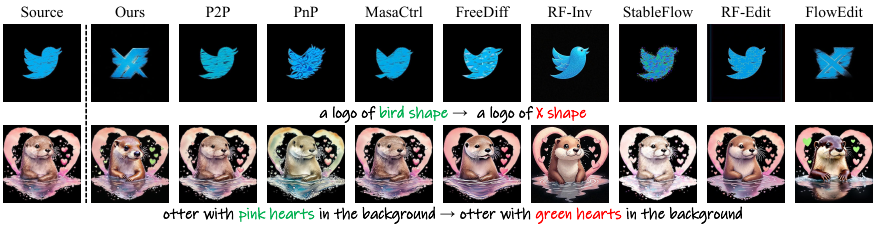}
\end{center}
\caption{
Challenging failure cases.
Examples where all methods fail to generate satisfactory edits. (Top) Large structural change from bird to X fails for all methods. (Bottom) Color-change edit where all methods miss the hollow heart in the background.
}
\label{fig:SM6-failure}
\end{figure*}

In this section, we present challenging cases where all methods, including ours, fail to produce satisfactory edits, as shown in Fig.~\ref{fig:SM6-failure}.
The first example involves a drastic structural transformation, changing a bird into an X-shaped object. All methods struggle with this task. While our method performs slightly better, the result still falls short of completing the intended transformation. Even methods like FlowEdit, which prioritize semantic alignment, fail here. This suggests that handling large-scale structural changes may require additional mechanisms, such as introducing controlled randomness or alternative guidance strategies.
The second example highlights a case where most methods fail to recognize a large hollow heart shape in the background. This is likely due to a mismatch between the visual content and the semantic prompt. Our method successfully turns all small solid hearts green but still misses the intended hollow shape. This points to a future challenge: developing editing techniques that can interpret and match abstract or contour-based semantics more effectively.

\begin{table*}[htbp]
  \centering
  \small
  \captionsetup{singlelinecheck=off, justification=raggedright}
  \begin{tblr}{
  colspec={c c c cccc cc},
  vline{2}={1-Z}{},
  vline{3,4}={1-Z}{dashed},
  hline{1,Z}={1pt},
  hline{2}={1-Z}{},
  hline{4}={1-Z}{dashed}
  }
  Method   &   ConvNeXt-T  & Aug & PnP &   MasaCtrl    &   FlexiEdit   &   FreeDiff  &   FlowEdit    &   Ours\\

  Distance$_{\times10^3}\downarrow$& - & - & $40.57$ & $36.18$ & $289.19$ & $21.33$ & $\underline{11.75}$ & $\mathbf{6.69}$ \\
  CLIP $\uparrow$   & - & - & $\underline{18.18}$ & $15.50$ & $17.64$ & $\mathbf{18.41}$ & $17.16$ & $16.51$\\

  AUC (\%) $\uparrow$ & $81.54$ & $82.10$ & $81.98 $& $\underline{84.22}$ & $82.05$ & $82.25$ & $83.83$ & $\mathbf{85.05}$ \\
  PR-AUC (\%) $\uparrow$    & $38.66$ & $38.81$ & $38.53$ & $38.82$ & $37.97$ & $38.60$ & $\underline{40.34}$ & $\mathbf{43.81}$ \\
  \end{tblr}
  \caption{Bleeding image editing results and classification performance after data augmentation. Aug represents traditional augmentation. \textbf{Bold}: best; \underline{underline}: second-best.}
  \label{tab:DownStreamApp}
\end{table*}

\section{More Details for Downstream Task}
\subsection{Task Overview}

Intraoperative hemorrhage recognition is a critical aspect of surgical workflow analysis, especially for early-stage bleeding events. Accurate recognition of bleeding during surgery helps surgeons maintain better control, improve procedural quality, and reduce potential risks. Typically, intraoperative bleeding is categorized into five levels~\cite{BloodDataset}: very low, low, intermediate, high, and very high, with increasing clinical risk.
To explore the practical potential of image editing techniques in real-world clinical tasks, we investigate the application in data augmentation for bleeding classification. Specifically, we use the newly released Laparoscopic Roux-en-Y Gastric Bypass dataset~\cite{BloodDataset}, which contains annotated images labeled by bleeding amount.

We apply our and other five editing methods to augment the dataset with synthetic hemorrhage images and evaluate whether these edited images improve model performance on a downstream classification task. The experiment follows two stages: we first train a classifier ConvNeXt-T~\cite{ConvNeXt} on the original dataset, then retrain it with the augmented data (\textit{i.e.}, edited images) to measure performance improvements.

\subsection{Experimental Setup and Evaluation Metrics}
The Laparoscopic Roux-en-Y Gastric Bypass dataset consists of 140 surgical videos, with 100 videos used for training and 40 for testing. Frames are sampled at 1 frame per second (fps), resulting in a total of approximately 770K frames. The training set contains 512K non-bleeding images and 44K bleeding images, while the test set contains 197K non-bleeding and 17K bleeding frames, reflecting an extreme class imbalance between bleeding and non-bleeding samples.

To evaluate the effectiveness of image editing for downstream clinical tasks, we follow the binary classification setup proposed in \cite{BloodDataset}, treating the problem as distinguishing bleeding from non-bleeding. Given the large data volume and computation constraints, we focus on a representative comparison of six editing methods to assess editing-based augmentation.
Specifically, we select 4,803 early-stage bleeding images from the 100 training videos (approximately 50 frames per video) and apply zero-shot editing using either Stable Diffusion v1.5~\cite{LDM1.5} or v3.5-Medium~\cite{SD35} pretrained on natural images. The editing prompts simulate different bleeding severity, \textit{e.g.}, converting minor bleeding into ``large amount of blood lost''. For baseline comparison, we also include a group with traditional data augmentation techniques such as rotation and flipping. The edited images are added back into the training set, and a ConvNeXt-T classifier pretrained on ImageNet is fine-tuned on the extended dataset.

Since bleeding is not spatially localized and edits affect global image appearance, we assess image editing quality using Structural Distance~\cite{StructureDistance} and CLIP similarity~\cite{CLIPsimilarity} for whole image, which measure content preservation and semantic alignment, respectively.
For classification performance, we report Accuracy, Precision, Recall, F1-score, AUC (Area Under ROC Curve), and PR-AUC (Area Under Precision-Recall Curve) to account for the data imbalance. Among these, PR-AUC is particularly important under imbalanced settings, which evaluates the trade-off between precision and recall across different classification thresholds. Unlike ROC-AUC, which may be misleading when negative samples dominate, PR-AUC more directly reflects the classifier’s ability to correctly identify the minority (bleeding) class.

\subsection{Quantitative and Qualitative Results}

\begin{figure*}[htbp]
\begin{center}
	\includegraphics[width=1\linewidth]{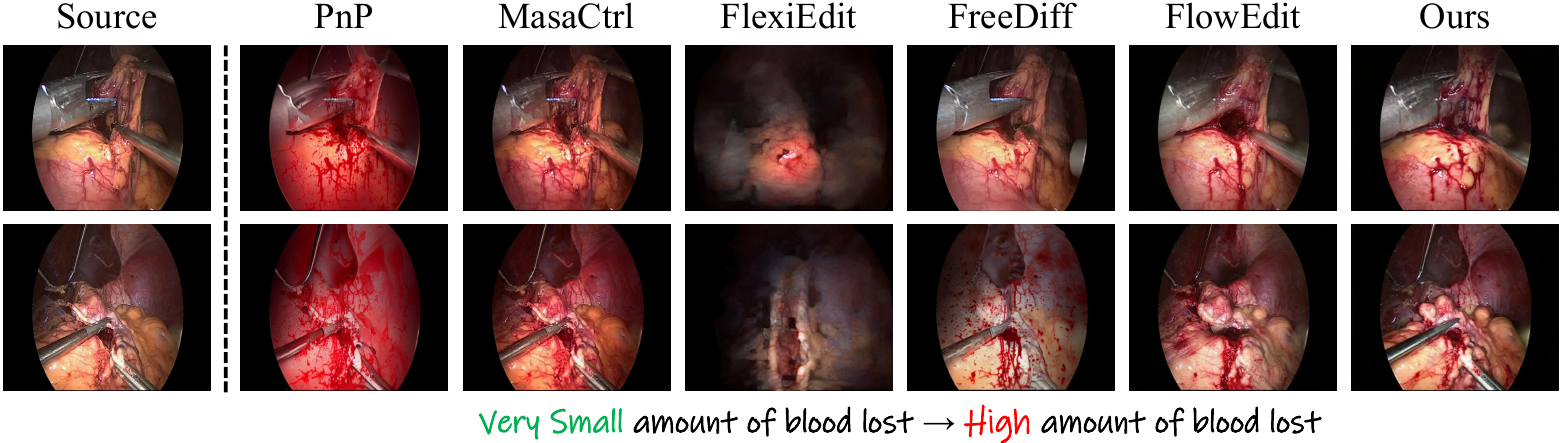}
\end{center}
\caption{
Qualitative results of bleeding image editing. Only Ours and FlowEdit achieve realistic and semantically meaningful edits. However, FlowEdit suffers from poor background preservation, with surgical instruments disappearing, whereas Ours maintains both semantic accuracy and contextual consistency.
}
\label{fig:SM7-blood}
\end{figure*}

Quantitative and qualitative results of bleeding image editing are shown in Table~\ref{tab:DownStreamApp} and Fig.~\ref{fig:SM7-blood}. For simplicity, only AUC and PR-AUC are reported here, while full classification results are included in the main paper. As shown in the visualizations, SD1.5-based methods struggle to generate realistic bleeding images. In contrast, FlowEdit and Ours produce more plausible and semantically aligned edits. However, FlowEdit often fails to preserve background details, such as missing surgical instruments in the second row.

Table~\ref{tab:DownStreamApp} demonstrates that our method achieves superior AUC and PR-AUC, indicating that high-quality, semantically faithful, and contextually consistent edits are more effective for augmenting bleeding datasets and improving downstream classification performance. Overall, this experiment validates the clinical potential of image editing techniques in surgical applications.
Future work may explore extending these methods to video-level editing, which could further enhance bleeding recognition in surgical videos.

\end{document}